\documentclass[12pt,letterpaper]{article}
\usepackage{mathptmx}
\usepackage{amsmath}
\usepackage{amssymb}
\usepackage{booktabs}
\usepackage[letterpaper]{geometry}
\usepackage[pdftex]{graphicx}
\usepackage{xcolor}
\usepackage{xurl}
\usepackage[english,colorlinks=true,allcolors=blue,breaklinks=true]{hyperref}

\usepackage{booktabs}
\usepackage{authblk}
\usepackage{doi}

\usepackage[numbers]{natbib}
\title{Entropic Hetero-Associative Memory}
\author[1]{Rafael Morales}
\author[2,*]{Luis A. Pineda}
\affil[1]{Universidad de Guadalajara, CUGDL, Guadalajara, 44130, Mexico}
\affil[2]{Universidad Nacional Aut\'onoma de M\'exico, IIMAS, Mexico City, 04510, Mexico}
\affil[*]{lpineda@unam.mx}

\begin{document}
\maketitle
\begin{abstract}
    \noindent The Entropic Associative Memory holds objects in a 2D relation or ``memory plane'' using a finite table as the medium. Memory objects are stored by reinforcing simultaneously the cells used by the cue, implementing a form of Hebb's learning rule. Stored objects are ``overlapped'' on the medium, hence the memory is indeterminate and has an entropy value at each state. The retrieval operation constructs an object from the cue and such indeterminate content. In this paper we present the extension to the hetero-associative case in which these properties are preserved. Pairs of hetero-associated objects, possibly of different domain and/or modalities, are held in a 4D relation. The memory retrieval operation selects a largely indeterminate 2D memory plane that is specific to the input cue; however, there is no cue left to retrieve an object from such latter plane. We propose three incremental methods to address such missing cue problem, which we call random, sample and test, and search and test. The model is assessed with composite recollections consisting of manuscripts digits and letters selected from the MNIST and the EMNIST corpora, respectively, such that cue digits retrieve their associated letters and vice versa. We show the memory performance and illustrate the memory retrieval operation using all three methods. The system shows promise for storing, recognizing and retrieving very large sets of object with very limited computing resources.
    
    \smallskip
    \noindent\textbf{Keywords:} hetero-associativity memory, declarative memory, indeterminacy of memory, memory and Hebb's learning rule.
\end{abstract}

\flushbottom


\section{Introduction}
\label{sec:introduction}

The Entropic Associative Memory (EAM) is a novel memory system in which recollections are stored diagrammatically in a table, or memory plane. The basic model defines three memory operations in relation to a cue: $\lambda$-register, $\eta$-recognition and $\beta$-retrieval. Memory objects are represented as mathematical functions such that columns and rows of the table correspond to function arguments and values, respectively, which in turn correspond to the object's attributes and values, as in standard feature-value structures \cite{internationalorganizationforstandardizationLanguageResourceManagement2006}. Functions representing diverse memory objects can ``overlap'' on the memory medium. The system has been tested with the storage, recognition and retrieval of manuscript digits \cite{pineda-eam-2021}, manuscript digits and letters \cite{morales-pineda-2022}, phonetic representations \cite{pineda-weam-2022}, and images of clothes, bags and shoes \cite{pineda-imagery-eam-2023}. 
 
From a computational perspective, the memory is declarative, abstractive, indeterminate and constructive. It supports the recovery not only of the cued objects but also of objects associated to the cue, as well as the production of association chains, possibly including objects of different classes\cite{pineda-imagery-eam-2023}, but the memory is mostly \emph{auto-associative}. A more general form of associations can be established between objects of different domains and modalities, and a memory supporting this functionality is \emph{hetero-associative}. Hetero-associative neural networks models were explicitly introduced with Kosko's BAM model \cite{bam}, which inspired a very large body of work  \cite{Bandy-Datta-1996,Kuruppu-1996,ritter-1999,yanez-2018,Kosko-2021}.

In this paper we present the Entropic Hetero-Associative Memory (EHAM) as an extension of EAM \cite{pineda-eam-2021,morales-pineda-2022,pineda-weam-2022,pineda-imagery-eam-2023} that opposes BAM and related models in the same ways that EAM oppose auto-associative neural network-based models, such as Hopfield's \cite{hopfield-1982}.

The structure of the paper is as follows: We present the extension of EAM to the hetero-associative model EHAM. The new model is tested with a heterogeneous corpus constituted by arbitrary associations between individual digits of MNIST and individual letters of EMNIST of designated associated classes. We illustrate the machinery with the object retrieved in both directions with three methods. Finally, we highlight the main features of the model, asses the results, and discuss some implications for research on associative memories.

\section{Hetero-Associative Extension of EAM}
\label{sec:EHAM}

\noindent The model allows the expression of pairs of objects, possible of different domains and modalities that can be registered and recognized as units, so one object of the pair can be used as cue to recover the other and vice versa. Let $A = \{a_1,...,a_n\}$ and $V = \{v_1,...,v_p\}$, $B = \{b_1,...,b_m\}$ and $ Z = \{z_1,...,z_q\}$, be two pairs of sets of attributes and values, respectively. An Entropic Hetero-Associative Memory $M_{A,B}$ stores associations between objects represented as functions $A\to V$ and $B\to Z$, respectively. The hetero-association between $f_a:A\to V$ and $f_b:B\to Z$ is represented by the function $f_{a,b}: A \times B \to V \times Z$ such that $((a_i,b_j),(v_k,z_l)) \in f_{a,b}$ if and only if $(a_i,v_k) \in f_a$ and $(b_j,z_l) \in f_b$. For simplicity, we use the notation $(a_i,b_j,v_k,z_l)$ to state the pair $((a_i,b_j),(v_k,z_l))$, so that relations in $A\times B\to V\times Z$ are expressed as relations in $A\times B\times V\times Z$.  The memory has three parameters, $\iota, \kappa, \xi\in\mathbb{R}^{+}$.

Let $M_{A,B}$ be a four-dimensional weighted relation $h\subset A\times B\times V\times Z$, and let the function $H:A\times B\times V\times Z\to W = \{0,\dots,w\}$, where $w$ is an integer larger than zero,  specify the weights of $h$ such that $(a_i,b_j,v_k,z_l)\in h$ if and only if $H(a_i,b_j,v_k,z_l) \ne 0$. Let $F_{a,b}$ specify the weights of $f_{a,b}$.

For each pair $(a_i, b_j)$ let $n_{i,j}$ be the number of nonzero values of the function $H_{i,j}: V\times Z\to W$, defined as $H_{i,j}(v_k, z_l) =H(a_i,b_j,v_k,z_l)$ and associated to the relation $h_{i,j}: V\to Z$ defined as $(v_k,z_l)\in h_{i,j}$ if and only if $(a_i, b_j, v_k, z_l)\in h$. Let $\omega_{i,j} = 0$ if $n_{i,j} = 0$ else
$$
    \omega_{i,j} = \frac{1}{n_{i,j}}\sum_{k = 1, l =1}^{m,q}H_{i,j}(v_k, z_l)
$$
the average nonzero weight in $H_{i,j}$. Then set $h_\iota \subset h$ to be the relation associated to the table $H_\iota$ defined as $H_\iota(a_i, b_j, v_k, z_l) = H(a_i, b_j, v_k, z_l)$ if $H(a_i, b_j, v_k, z_l) > \iota\omega_{i,j}$ else $H_\iota(a_i, b_j, v_k, z_l) = 0$. Let
$$
    \omega_H = \frac{\sum_{i=1,j=1}^{n, p}\omega_{i,j}}{nm}.
$$

Let $r\subset A\times B\times V\times Z$ be a weighted relation serving as cue, and be $R$ its weight function. Let $s_R$ be the sum of all $R$ values. Then let $R' = R$ if $s_R = 0$ else $R' = R/s_R$. Initially, $M_{A,B}$ is empty, so $h = \emptyset$ and $H = 0$. The memory operations are defined as follows:
\begin{itemize}
	\item Memory Register: $\lambda(h,f_{a,b}) = q$ such that $q = h\cup f_{a,b}$ and its weight function $Q(x) = H(x) + F_{a,b}(x)$ if $H(x) + F_{a,b}(x) \le w$, else $Q(x)= w$, for all $x\in A\times B\times V\times Z$.
    \item Memory Recognition: $\eta(r,h)$ is true if and only if
    \begin{itemize}
        \item $R(x) \implies H_\iota(x)$ (material implication, considering as false a value of zero, and true otherwise) for all $x \in S \subset A\times B\times V\times Z$ such that $nmpq - |S| \le \xi$, and
        \item being $H'$ the cell by cell multiplication of $H$ and $R'$, and $\rho$ the sum of all cells in $H'$, $\rho \ge \kappa\omega_H$, 
    \end{itemize}
    \item Memory Retrieval:
        \begin{itemize}
          \item $\beta_{b}(f_a, h)=f_b$ such that:
            \begin{enumerate}
             \item Let $F_a$ be the weight function of $f_a$, $s_F$ the sum of all its weights, and let $F_a' = F_a$ if $s_F = 0$ else $F_a' = F_a/s_F$. 
             \item For each $(a_i,v_k) \in f_a$ the weighted relation $h_{a_i,v_k}: B \rightarrow Z$ is recovered such that $h_{a_i,v_k}=\{(b_j,z_l)|(a_i, b_j, v_k, z_l) \in h\}$, and its corresponding weight function $H_{a_i,v_k}: B \times Z \rightarrow W$, such that $H_{a_i,v_k}(b_j,z_l)= F_a'(a_i)H(a_i, b_j, v_k, z_l)$.
             
             \item Let $h_{f_a} = \bigcup_{(a_i, v_k)\in f_a}h_{a_i,v_k}$ be a weighted relation in $B\times Z$, and  $H_{f_a}= \sum_{(a_i,v_k) \in f_a}H_{a_i,v_k}$ its weight function.

            \item Let $\Omega_j$ be a probability distribution over $Z$ defined by the weights in the table $H_{f_a}$ located in the column corresponding to $b_j$. Select $f_b(b_j)$ randomly from $Z$ using $\Omega_j$ for all $b_j\in B$.
         \end{enumerate}
         \item $\beta_a(f_b, h) = f_a$ is defined by exchanging the roles of $f_a$ and $f_b$, $A$ and $B$, and $V$ and $Z$, in the definition above.
         \end{itemize}
\end{itemize}
The memory has an entropy at any state. The entropy is a direct extension of the W-EAM \cite{pineda-weam-2022}. Let
$$
    W_{i,j} = \sum_{k = 1, l =1}^{m,q}H_{i,j}(v_k, z_l)
$$
be the accumulated weight for the pair $(a_i,b_j)$ in $A \times B$ --i.e., the sum of its corresponding cells in the co-domain $V \times Z$. The probability $p_{i,j,k,l}$ of the cell $h(a_i,b_j,v_k,z_l)$ is
$$
p_{i,j,k,l} =
\begin{cases}
H(a_i,b_j,v_k,z_l)/W_{i,j}&\text{if } W_{i,j} \ne 0\\
0&\text{otherwise.}
\end{cases}
$$
The entropy of every pair $(a_i,b_j)$ in the domain $A \times B$ is Shannon's entropy, as follows:
$$
e_{i,j}= -\sum_{k=1,l=1}^{m,q} p_{i,j,k,l}\log_2(p_{i,j,k,l})\enspace\mbox{where} \enspace \log_2(p_{i,j,k,l}) = 0\enspace\mbox{if}\enspace p_{i,j,k,l} = 0.
$$
The entropy of the whole relation $h$ is the average entropy of all objects in the domain $A \times B$:
$$
e_h = \frac{1}{nm}\sum_{i=1,j=1}^{n,m}e_{i,j}.
$$

\section{Experiments with the HEMNIST Corpus}
\label{sec:Experiments}

\noindent We created two auto-associative models using the weighted EAM system \cite{pineda-imagery-eam-2023} for the MNIST \cite{lecunMNISTDatabaseHandwriting} dataset of digits, and a selection of then classes from the EMNIST Balanced dataset of letters \cite{cohen2017emnist}. The memory system as a whole involves three levels of representation:
\begin{enumerate}
    \item A local concrete modality specific level sustaining the input and output images.
    \item A local abstract amodal level in which images in the first level are represented as functions.
    \item The distributed abstract amodal level which constitutes the memory proper.
\end{enumerate}

\begin{figure}[ht]
\includegraphics[width=0.8\textwidth]{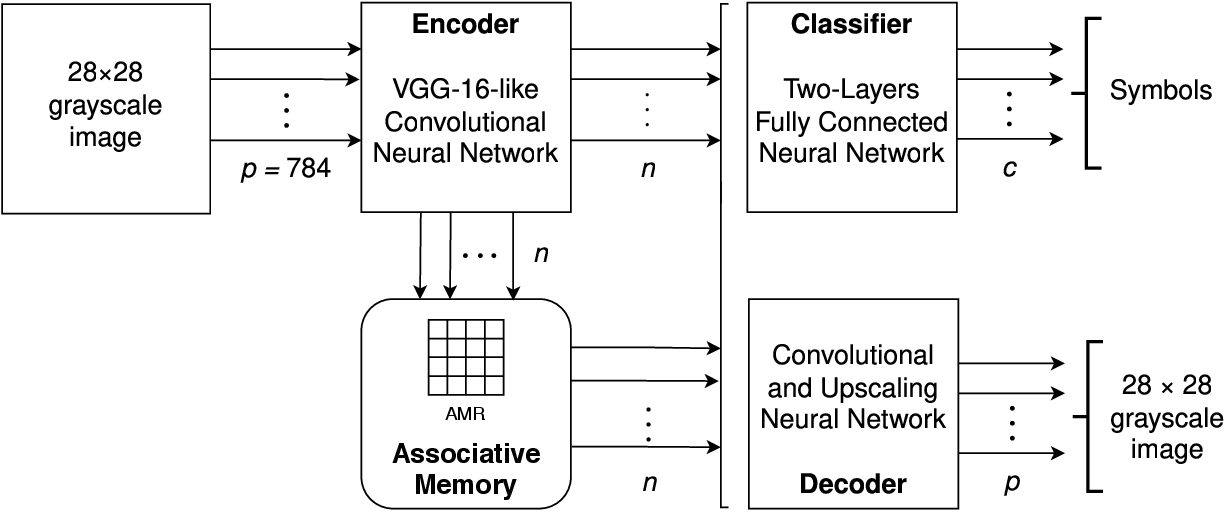}
\centering
\caption{System Architecture. Source: \protect\cite{pineda-imagery-eam-2023}.}
\label{fig:WEAM-architecture}
\end{figure}

The architecture uses an autoencoder and classifier neural network \cite{pineda-imagery-eam-2023} for the conversion between representational levels and for classifying the objects in their abstract representations. The architecture is illustrated in Figure \ref{fig:WEAM-architecture}, as was used in this and in our previous work \cite{pineda-imagery-eam-2023}.

\begin{figure}[htbp]
\includegraphics[width=0.7\textwidth]{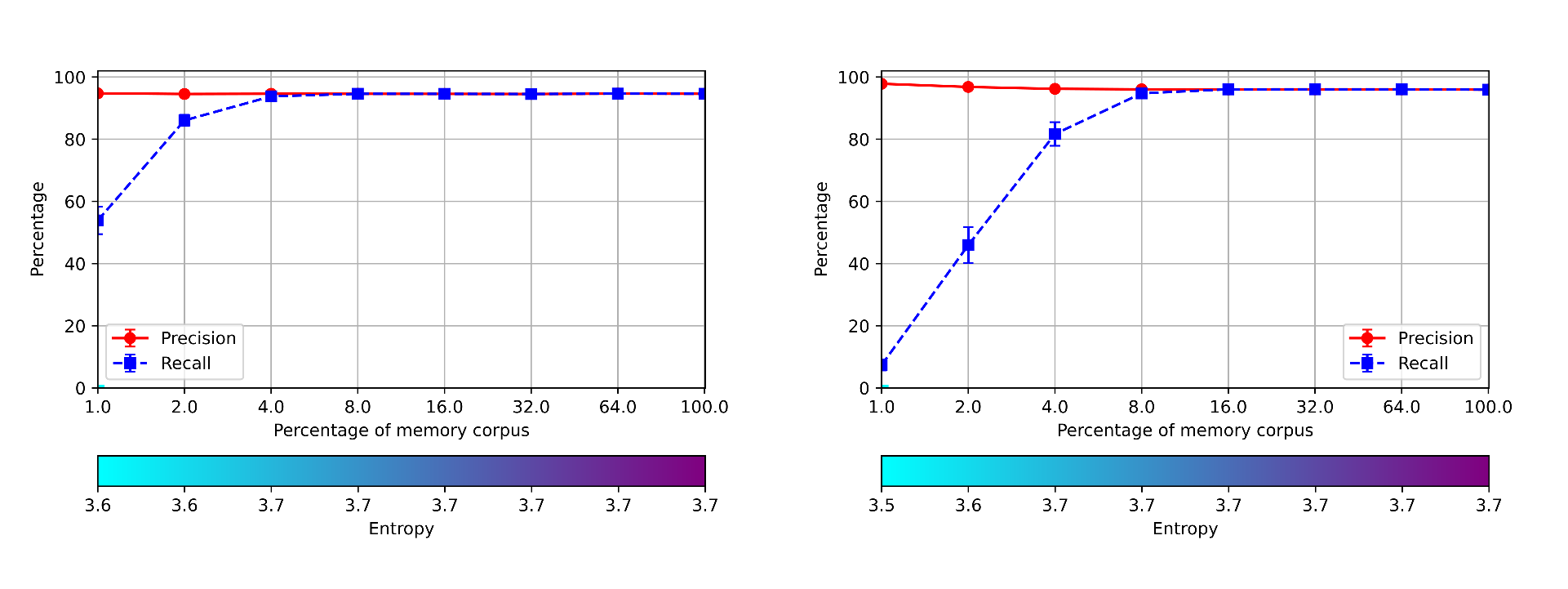}
\centering
\caption{Performance achieved in ten-fold cross validation by EAM for MNIST (left) and a subset of EMNIST Balanced (Table~\ref{tab:classes-HEMNIST}) (right) using increasing percentages of the filling corpus. The parameters are set to their default values of $\iota=0$, $\kappa=0$ and $\xi=0$.}
\label{fig:performance-MNIST-EMNIST}
\end{figure}

For the present experiments we tested tables of $32 \times 16$, $64 \times 16$ and $128 \times 16$, for storing objects in MNIST \cite{lecunMNISTDatabaseHandwriting} and the subset of ten classes (Table~\ref{tab:classes-HEMNIST}) of EMNIST Balanced \cite{cohen2017emnist} datasets, being the second configuration the best in terms of the performances of both the neural networks and both associative memory registers (AMRs). The performances of the neural networks are presented in Table~\ref{tab:nnets-performance}, while the performances of the auto-associative memories are presented in Figure~\ref{fig:performance-MNIST-EMNIST}.

\begin{table}[htbp]
    \centering
    \caption{Performances achieved in ten-fold cross validation by the neural networks (autoencoder plus classifier) for MNIST and a subset of EMNIST Balanced (Table~\ref{tab:classes-HEMNIST}), with an interface of 64 values between the encoder (output) and the decoder and classifer (inputs).}
    \begin{tabular}{@{}lrrrrrrrrrrrr@{}}
        \toprule
        Dataset&\multicolumn{4}{c}{Classifier accuracy (\%)}&\multicolumn{4}{c}{Autoencoder RMSE (\%)}\\
        &Min.&Mean&Max.&SD&Min.&Mean&Max.&SD\\
        \midrule
         MNIST &95.4&96.2&96.7&0.4&10.2&10.4&10.5&0.12\\
         EMNIST Balanced&96.6&97.1&97.8&0.4&12.0&12.2&12.5&0.16\\
         \bottomrule
    \end{tabular}
    \label{tab:nnets-performance}
\end{table}

For testing the new machinery we created the HEMNIST corpus constituted by pairs formed with randomly selected digits and letters from designated classes of MNIST and EMNIST. Table \ref{tab:classes-HEMNIST} shows the corresponding classes of the two basic corpus. The corpus size is restricted by the number of samples per class available in EMNIST Balanced, whose average is 2800. For the experiments the corpus was divided into three mutually exclusive partitions, as follows:
\begin{itemize}
    \item 70\% for training the autoencoder and the classifier (in turn divided into 80\% for training and 20\% for validating);
    \item 20\% for filling the AMR (i.e., remembering corpus); the number of objects in this portion is about 5600, so there are about 560 pairs per association class.
    \item 10\% for testing --i.e., about 2800 object or 280 pairs for each of the ten associated classes.
\end{itemize}

\begin{table}[htbp]
    \centering
    \caption{Related classes of MNIST and EMNIST Balanced for the construction of HEMNIST}
    \begin{tabular}{@{}lcccccccccc@{}}
        \toprule
        Dataset&\multicolumn{10}{c}{Classes (symbols)}\\
        \midrule
         MNIST & 0 & 1 & 2 & 3 & 4 & 5 & 6 & 7 & 8 & 9 \\
         EMNIST & T & O & P & B & W & Z & M & A & L & S\\
         \bottomrule
    \end{tabular}
    \label{tab:classes-HEMNIST}
\end{table}

As objects both in MNIST and EMNIST are represented by functions with 64 arguments and 16 possible values, the dimensions of the hetero associative memory register (HAMR) used for all the experiments are $64 \times 64 \times 16\times 16$. 

The experiments were verified with a standard 10-fold cross validation process, and the results bellow show the average of the 10 partitions in all experimental situations.

\subsection{Recognition experiments}

\begin{figure}[htbp]
\includegraphics[width=0.4\textwidth]{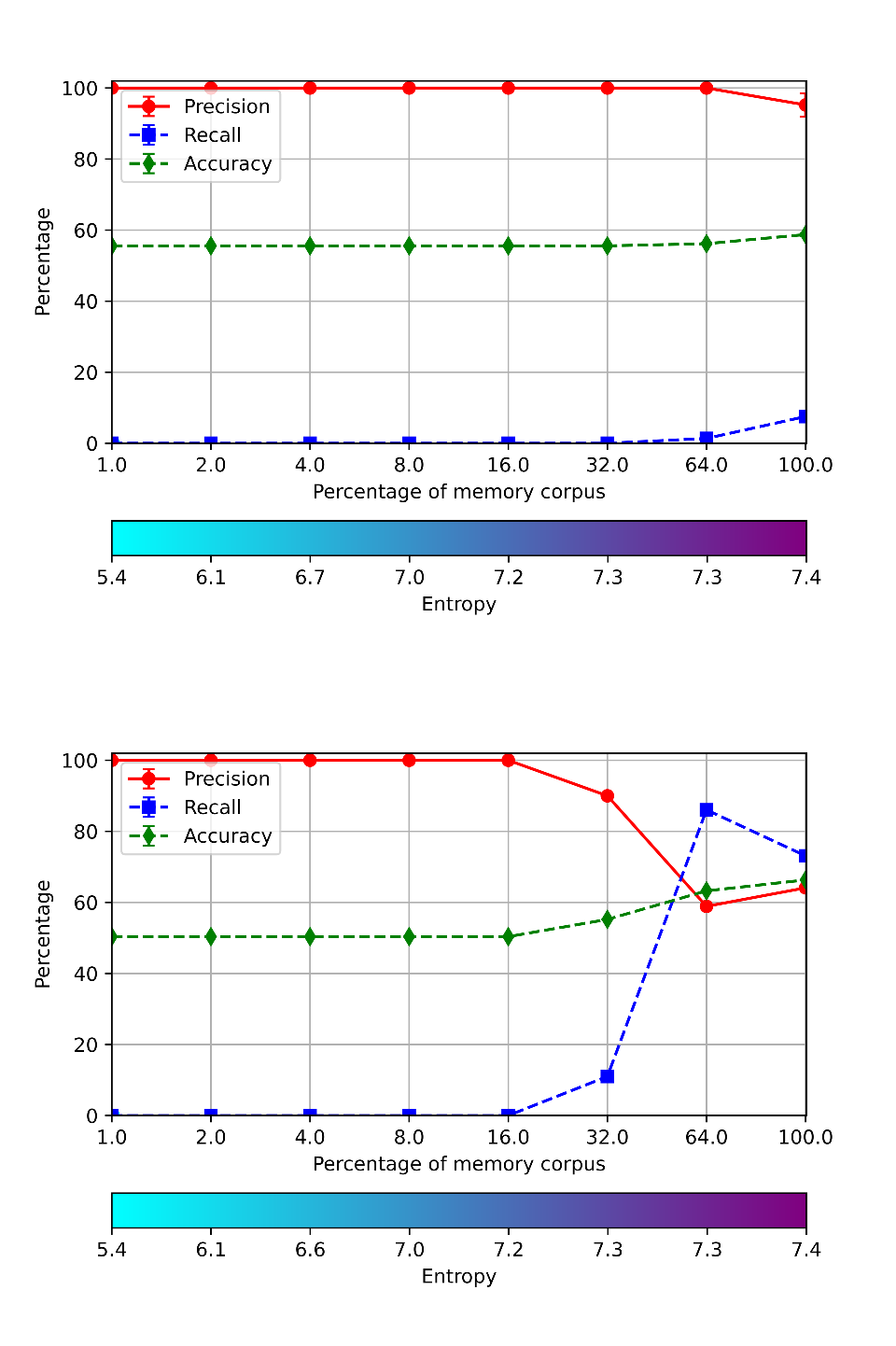}
\centering
\caption{Recognition precision, recall and accuracy setting the recognition parameters to their default value --$\iota=0$, $\kappa=0$ and $\xi=0$-- (top). Recognition performance using $\iota=0.05$, $\kappa=32$ and $\xi=0$ (bottom).}
\label{fig:hetero-recognition-performance}
\end{figure}

We constructed a recognition tests constituted as follows:

\begin{itemize}
    \item 50\% by pairs included in the test corpus of HEMNIST, balanced for each pair of classes;
    \item 50\% by pairs not included in HEMNIST, formed randomly out an object of MNIST and an object of EMNIST selected from classes that do not correspond in HEMNIST --see Table~\ref{tab:classes-HEMNIST}.
\end{itemize}
The recognition performance is measured in terms of the precision, recall and accuracy in relation to the percentage of the remembering corpus included in the memory. The recognition performance of the EHAM is shown in Figure 
\ref{fig:hetero-recognition-performance}.

The figure on top shows that, if the parameters are set to their default values, the precision rate is in general quite high but the recall is very low, suggesting that the system is not yet within its operational range. For this, we relaxed the $\eta$-recognition operation by setting the parameter $\xi=32$, which relaxes the recognition test by less than 1\% of the total number of arguments. We also increased slightly the value of $\iota$ to 0.05. This parameter's setting makes the memory system operational, and the best trade-off between precision and recall occurs when the totality of the corpus is included, where the precision and the recall are around 65\% and 70\%, respectively. The tendency in both graphs suggests the number of objects registered can be increased significantly. 

\subsection{Retrieval experiments}
\label{sec:recalling-experiments}

\noindent The $\beta$-retrieval operation in relation to a cue in the source memory field reduces the hetero-associative memory 4D relation to a 2D relation, that is specific to the cue, in the target memory field, from which the remembered object has to be produced, as in the basic EAM model. However, there is no cue available in the target field to complete the operation, giving rise to \emph{the missing cue problem}. Next we present three incremental strategies to address such problem with their corresponding results.

\subsubsection{Random samples method (RS)}
\label{sec:mcp-random}

The basic strategy consists on the random selection of the retrieved object, as presented in Section \ref{sec:EHAM} in step 4 of the definition of the $\beta$-retrieval operation. Figure \ref{fig:mcp-solution} (top) shows the precision and recall of the memory retrieval operation, in both directions --i.e., from MNIST to EMNIST and vice versa-- in relation to the amount of the remembering corpus included in the memory registers. 

The memory becomes operational when 16\% of the remembering corpus is included. The best compromise, in both directions, occurs when the memory is filled with 32\% of the corpus, with precision of 43.3\% and recall of 40.3\%, from MNIST to EMNIST, and precision of 35.2\% and recall of 33.9\% in the reverse direction. Although the performance is low, this experiment shows that the memory plane in the target field, produced out of a cue in the source field, does include the object to be retrieved, with a probability that is greater than chance.

\subsubsection{Sample and test method (ST)}
\label{sec:mcp-sample-test}

We now consider that the image back in the source field of the object retrieved in the target one must match the original cue to the EHAM $\beta$-retrieval operation. This observation suggests a sample and test method to generate and select an object in the target field that is hetero-associated to the original cue in the source field, as follows:

\begin{enumerate}
    \item Select the relation $r_T$ in the target field $T$ on the basis of cue $c_S$ in source field $S$.
    \item Select a number of objects $o_T$ randomly from $r_T$.
    \item For each $o_T$ produce its backwards relation $r_S$ in the source field $S$ and assess its similarity to $c_S$.
    \item Select the object $o_T$ that produces the $r_S$ which is the most similar to $c_S$.
\end{enumerate}

To asses the similarity between a relation and a cue we use the following method: let the sets $A = \{a_1,...,a_n\}$ and $V = \{v_1,...,v_m\}$ be the domain and codomain of a weighted relation $r: A\to V$ stored in memory, and let the function $R: A \times V \to  \{0,..., l\}$, where $l$ is an integer greater than zero, specify the weights of $r$. Now, let $f\,$ be a weighted function with the same domain and codomain as $r$ representing a cue, with $F: A \to  \{0,..., l\}$ as its weights. The distance $d$ between $r$ and $f$ is defined as follows: if $\eta(r,f)$ is false $d(r, f) = \infty$, otherwise
\begin{displaymath}
	d(r,f) = \frac{\sum_{i=1}^{n}F(a_i)d_i}{\sum_{i=1}^{n}F(a_i)}\quad
	\mbox{where}\quad d_i = \frac{\sum_{j=1}^{m}R(a_i,v_j)(v_j - f(a_i))^2}{\sum_{j=1}^{m}R(a_i,v_j)}.
\end{displaymath}

\begin{figure}[htbp]
\includegraphics[width=0.7\textwidth]{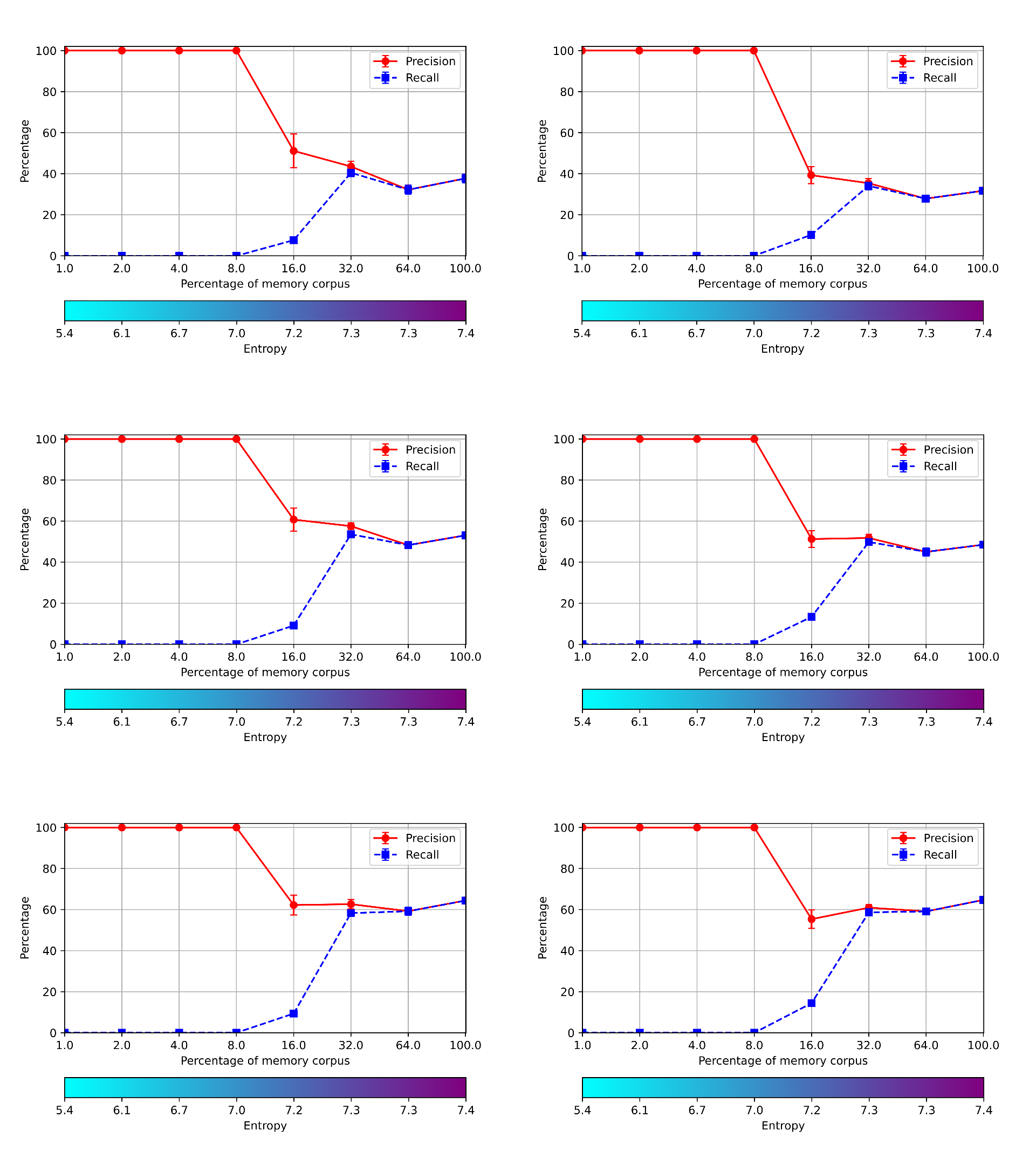}
\centering
\caption{Performance of the random method (top); the sample and test method (middle) and the sample and search method (bottom). From a digit of MNIST produce a letter of the EMNIST (left), and vice versa (right).}
\label{fig:mcp-solution}
\end{figure}

Figure \ref{fig:mcp-solution} (middle) shows the precision and recall using this strategy, with a number of samples of 128. As can be seen, the sample and test method shows the best trade-off between precision and recall at around 55\% from MNIST to EMNIST and 50\% in the reverse direction, when 32\% of the remembering corpus is included.

\subsubsection{Sample and search method (SS)}
\label{sec:mcp-sample-search}

We augment the method in \ref{sec:mcp-sample-test} with a local search process (random descent) \cite{hasleDiscreteOptimizationHeuristics2009} that optimizes the objects selected by sampling and test, so the distance of the backwards relation $r_S$ of each $o_T$ to $cue_S$ is minimized. Intuitively, the process consists on exploring the neighborhood of each $o_T$ and modifying it monotonically reducing such distance at each step.

Figure \ref{fig:mcp-solution} (bottom) shows the results of the sample and search method. The search process produces an average of 800 objects per $o_T$ delivered by sampling and test; hence, this method produces around $128 + 800 = 928$ candidate instances from which the one whose backwards image approximates better the original cue is chosen. As can be seen, the best trade-off between precision and recall is close to 60\% in both directions, when the full remembering corpus is included.

\section{Illustration of the machinery}
\label{sec:recollection-chains}

\begin{figure}[htbp]
\includegraphics[width=0.8\textwidth]{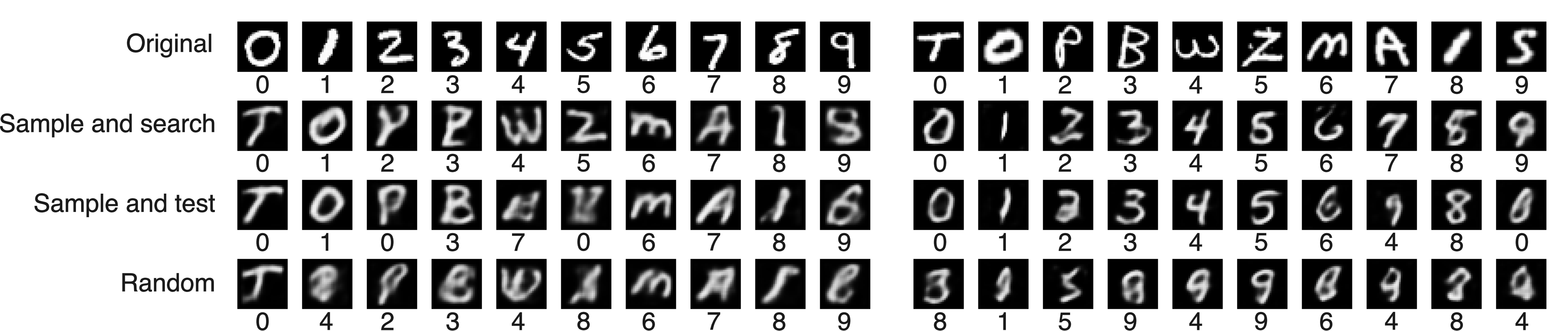}
\centering
\caption{Hetero-Associative recollection: letters from EMNIST retrieved using their associated digits of MNIST as cues (left) and digits from MNIST retrieved using their associated letters of EMNIST as cues (right). The top-row shows the initial cue, and the remaining rows show the objects produced by the $\beta$-retrieval operation using the sample and search, the sample and test, and the random methods, respectively.}
\label{fig:illustration-examples}
\end{figure}

Figure \ref{fig:illustration-examples} shows chosen examples from a random sample of 5\% the products of the $\beta$-retrieval operation for all classes using the three methods. The numbers below the symbols indicate the class selected by the classifier --see Figure \ref{fig:WEAM-architecture}. The top row shows the cue of $\beta$-retrieval for an instance digit and an instance letter of all ten classes, and the corresponding symbols produced by sample and search, sample and test and random, are shown in the second, third and fourth rows, respectively. The symbols produced by the sample and search and the sample and test methods have a reasonable good quality, and the random method has a lower quality, which is consistent with their corresponding classification rates.

\section{Results}
\label{sec:results}

We have introduced the Entropic Hetero-Associative Memory (EHAM) as a direct extension of the EAM model. We tested the model with memory recognition and memory retrieval experiments using hetero-associations between specific digits and letters taken from the MNIST and EMNIST corpora. The performance of EHAM in the recognition task, using both correct and incorrect associations as testing corpus, is moderately good (precision and recall around 65\% and 70\%, respectively). The memory retrieval operation, on its part, requires two memory cues: one for the source and one for the target memory plane, but the latter cue is missing, augmenting the indeterminacy of the problem significantly. We explore three incremental methods for completing the retrieval operation despite that the cue to the target plane is lacking. The results are shown and can be compared in Table \ref{tab:results}. The increase of performance reflects the reduction of the indeterminacy and the consequent increase of the precision and the recall for a given amount of remembering corpus, although at the expense of the computational demands of physical memory and computing time. The two first methods are fully declarative and the third  uses a very moderate amount of search --i.e., as compared to current generative AI system. The first is significantly better than random; the seconds improves considerable over the first; and the third improves a little over the second. We leave for further work whether there are other methods that can improve the performance of the retrieval operation.

\begin{table}[th]
    \centering
    \caption{Indeterminacy reduction and results. The \emph{time} column refers to the average time per fold in both directions taken by the cross-validation process.}
    \begin{tabular}{@{}cccccccc@{}}
        \toprule
         &&&\multicolumn{2}{c}{MNIST → EMNIST}&\multicolumn{2}{c}{EMNIST → MNIST}\\
         \cmidrule(lr){4-5}\cmidrule(rl){6-7}
         Method & \% of corpus & Samples & Precision & Recall & Precision & Recall & Time \\
         \midrule
         RS & 32 & 1 &  0.44 & 0.40 & 0.35 & 0.34 & 177.5 sec \\
         ST & 32 & 128 & 0.58 & 0.53 & 0.52 & 0.50  & 1.9 hrs\\
         SS & 100 & 128 & 0.59 & 0.59 & 0.60 & 0.60 & 12.8 hrs \\
         \bottomrule
    \end{tabular}
    \label{tab:results}
\end{table}

\section{Discussion}
\label{sec:discussion}

The key property of the EAM model is that the memory register operation ``overlaps'' memory cues on the medium --so cells may be used in the representation of different objects and, conversely, the represented objects may share different cells-- making the representation genuinely distributed. This operation models a form of Hebb's learning rule, as the simultaneous reinforcement of cells is causal to such cells being involved in the same computational function --e.g., remembering the particular object. Hence, the bindings between memory objects depend only on the weights of the cells used by their representations. The distributed property also gives rise to a huge memory capacity, and the memory operations can be implemented in parallel if the appropriate hardware is provided, in the spirit of the original presentation of the Connectionist program \cite{Rumelhart}. However, the identity of the stored cues is dissolved, the representation becomes indeterminate and has an entropy value, and the memory is operational in an entropy interval, which is nor too low neither too high. The distributive property gives rise as well to a very large set of potential objects that may allow for the recognition of novel views of known objects and imagination \cite{pineda-imagery-eam-2023} but also to false recollections \cite{MCDERMOTT-1996}. The memory retrieval operation recovers a novel object out of the indeterminate memory mass and the cue, and is genuinely constructive. All of these properties transfer from EAM to EHAM. In particular, the bindings between hetero-associated objects depend only on the weights of their representations in the hetero-associated memory planes.

The extension to the hetero-associative case revealed the missing cue problem which had not been previously observed. In standard settings, such as Kosko's BAM \cite{bam} and related work, the bindings between memory objects are concrete, contrasting with the present model in which a cue in the source field determines a largely indeterminate memory plane in the target field, where the sought object is held, but there is not a cue available for recovering such object. The effort to diminish the indeterminacy with the present three, and possible other methods, contributes to solve the problem in a context independent manner. We leave for further work whether memory retrieval may be thought of as a Bayesian inference such the object recovered ``within the system'' is considered a likelihood but there is also an external cue available, provided through the cognitive architecture, which is considered a prior. More generally, the EHAM model and the missing cue problem suggests that remembering ``in context'' does involve a cue that is specific to the particular act, but also an additional a priori cue, provided by the context. This mechanism may support Bartlett's emphatic observation that remembering is a conventional and rationalized task \cite{bartlett-1932}. The EHAM may also shed light on how remembering can be manipulated with strategies such as priming and perhaps the use of mnemonics.

Large computing system, such as current social and service robots, and related kind of computing agents do not support a long term memory of their own interactions with the world, and are very limited to store and make a profitable use of empirical knowledge collected by themselves. Likewise, generative AI programs --based on large language models-- implement a form of semantic memory, but their interactions take place in an a priori context, and there is no sense in which such systems remember episodic experiences and that their ``knowledge'' is grounded in the world. These limitations contrast greatly with agents endowed with a natural declarative memory, which are able to store and use a huge amount of recollections in relation to a cue from empirical experience, and from their own thought processes, using very limited computational resources; for this, the construction of large operative EHAM systems may impact in the development of computational declarative long-term learning.

\section{Experimental Setting}
\label{sec:setting}
The experiments were programmed in Python 3.11 on the Anaconda distribution. The neural networks were implemented with TensorFlow 2.14.0, and the graphs were produced using Matplotlib, ImageMagik, and Inkscape. The experiments were run on an Alienware Aurora R5 with an Intel Core i7-6700 Processor, 16 GBytes of RAM and an NVIDIA GeForce GTX 1080 graphics card.

\section{Data Availability}
\label{sec:data}
The datasets used in the present study are MNIST \cite{lecunMNISTDatabaseHandwriting} and EMNIST Balanced \cite{cohen2017emnist}. The full code and the detailed experimental results are available in Github at\\
\url{https://github.com/eam-experiments/hetero}.

\section{Acknowledgments}

The second author acknowledges the partial support of grants PAPIIT-UNAM IN112819 and IN103524, México. 

\bibliography{eham}

\end{document}